\documentclass[11pt,a4paper]{article}
\usepackage[hyperref]{acl2020}
\usepackage{times}
\usepackage{latexsym}
\usepackage{amsmath}
\usepackage{amsfonts}
\usepackage{multirow}
\usepackage{makecell}
\usepackage{dblfloatfix}
\usepackage{booktabs}

\usepackage{graphicx}
\usepackage{enumitem}
\usepackage{microtype}

\aclfinalcopy 
\def\aclpaperid{063} 

\title{Self-Supervised Claim Identification for Automated Fact Checking}

\author{Archita Pathak \\
  University at Buffalo (SUNY) \\
  Buffalo, NY \\
  \texttt{architap@buffalo.edu} \\\And
  Mohammad Abuzar Shaikh \\
  University at Buffalo (SUNY) \\
  Buffalo, NY \\
  \texttt{mshaikh2@buffalo.edu} \\\And
  Rohini K. Srihari \\
  University at Buffalo (SUNY) \\
  Buffalo, NY \\
  \texttt{rohini@buffalo.edu} \\}

\date{}

\begin{document}
\maketitle
\begin{abstract}
We propose a novel, attention-based self-supervised approach to identify ``claim-worthy'' sentences in a fake news article, an important first step in automated fact-checking. We leverage \textit{aboutness} of headline and content using attention mechanism for this task. The identified claims can be used for downstream task of claim verification for which we are releasing a benchmark dataset of manually selected compelling articles with veracity labels and associated evidence. This work goes beyond stylistic analysis to identifying content that influences reader belief. Experiments with three datasets show the strength of our model\footnote{Data and code available at: https://github.com/architapathak/Self-Supervised-ClaimIdentification}.
\end{abstract}



\section{Introduction}
\label{sec:intro}
{
The explosion of fake news on social media has resulted in global unrest and has been a major concern for governments and societies worldwide\footnote{https://www.reuters.com/article/us-singapore-politics-fakenews-factbox/factbox-fake-news-laws-around-the-world-idUSKCN1RE0XN}. According to a recent Pew Research Study, Americans rate it as a larger problem than racism, climate change, or illegal immigration\footnote{https://www.journalism.org/2019/06/05/many-americans-say-made-up-news-is-a-critical-problem-that-needs-to-be-fixed/}. Since, it's inexpensive to create a website and easily disseminate content on the social media platforms, there is a rising need for automated fake news detection. Furthermore, AI solutions are also required to follow good practices, specifically avoiding censorship, violation of fundamental rights such as freedom of expression, and ensuring data privacy \cite{de2018multi}. However, to date, AI models proposed for fake news detection do not scale for detecting real-time fake news\footnote{https://www.technologyreview.com/s/612236/even-the-best-ai-for-spotting-fake-news-is-still-terrible/}. 
} \newline
{
\indent Much of the research on automated text-based fake news detection can be classified into three broad categories: (1) linguistic approach, which focuses on lexical, stylometric and pattern learning mechanisms \cite{potthast2017stylometric,rashkin2017truth, wang2017liar,singhania20173han,perez-rosas-etal-2018-automatic}; (2) network-based approach, which leverages features such as the speed and volume of propagation of fake news articles on social media platforms \cite{castillo2011information,yang2012automatic,kwon2013prominent,Ma:2015:DRU:2806416.2806607,jin2016news,Ruchansky:2017:CHD:3132847.3132877, Wu:2018:TFF:3159652.3159677}; and (3) automated fact-checking approach, which is an effort to assist manual fact-checkers by automating some of their tasks such as detection and verification of claims \cite{graves2018understanding}.
}

{
\indent While most work in automated fact-checking has been focused on claim verification task, very few methods have been proposed for detection of claims \cite{hassan2017toward,jaradat-etal-2018-claimrank, konstantinovskiy2018towards}. The approaches in these efforts are majorly related to \textit{political discourse}. However, our focus is on \textit{fake news}, which are broader than \textit{political discourse} since (i) they are deliberately written with a divisive agenda to cause social unrest, (ii) they are not constrained to only politics, and (iii) the headline plays an equally important role in compelling people to read the article. 
}

{
In this paper, we focus on articles where there is a deliberate intent to influence readers through fabricated or manipulated claims in the headline and the content. Such articles have a compelling writing style similar to the mainstream media. Hence, we build datasets containing these type of compelling articles along with veracity labels and associated evidence supporting the label of each article. We, then, use these datasets to identify ``claim-worthy'' sentences. In our work, we define ``claim'' as \textit{``statements which are important to the point of the article but one would require to have them verified.''}} 

{Our working hypothesis is that in fake news which are created to cause harm, these are the sentences most relevant to the headline. Exploiting the hypothesis that the essence of a news article is encapsulated in its headline \citep{article, Kuiken, Wahl-Jorgensen}, we propose a self-supervised method to explore the \textit{aboutness} of the content with the headline of the article to extract the most relevant sentences. \citet{bruza1996study} defines \textit{aboutness} as: \textit{an information carrier i will be said to be \textbf{about} information carrier j if the information borne by j holds in i}. The idea is taken from Information Retrieval domain where it is used to signify implications between query and document, specifically to explore the underlying meaning or concept within the document and the query \cite{azzopardi2009advances}. In our work, headline is modelled as a query while each of the sentences of the article acts as a document, and we use the concept of \textit{aboutness} to find the relevant sentences. We show that attention-based mechanisms are able to successfully capture this concept in the news article.}

{
}

{
\textbf{Contribution:}  
In this work: (i) we introduce a self-supervised representation learning model that eliminates the prerequisite that requires human to annotate data, which is a time consuming and costly task; (ii) the proposed headline-to-sentence attention-based approach for claim identification is novel; previous unsupervised approach for this task use weak supervisory signal which does not capture the context of the article efficiently; and (iii) we propose a benchmark dataset for evidence-based fake news detection. Our dataset contains evidence for each of the fake news articles that contributes to the overall degree of veracity of the article.
}

\section{Related Work}
\label{sec:related}
{\textbf{Claim Identification/Detection:} The task of claim identification/detection was first introduced by \citet{levy_context_2014} who, with the help of human annotators, provided a dataset and a fundamental approach in identifying context dependent claims. In their dataset, which was originally developed by \citet{aharoni_benchmark_2014}, each statement indicates whether it should be considered as a context dependent claim (CDC) or not. \citet{levy_context_2014} reported encouraging results obtained through a supervised learning algorithm using a cascade of classifiers. A rule-based model was introduced by \citet{eckle-kohler_role_2015} to bifurcate claim and premise statements in an argumentative discourse environment. However, these methods were generic to only a small set of corpora. Furthermore, \citet{levy_unsupervised_2017} also introduced an unsupervised approach to detect claims, which involves a weak supervisory signal ``that'' for training. However, this approach does not capture the \textit{aboutness} of the article to understand the context of ``claim-worthy" sentences.
}

{
In 2017, \citet{hassan2017toward} introduced ClaimBuster, a platform developed by training a supervised learning model on a large annotated corpus of televised debates in the USA. Their model used SVM classifier to detect \textit{claim-worthy factual claims} and produced a score of how important a claim is to factcheck. The 20,000 sentences in the corpus were annotated by human coders to distinguish between claim-worthy factual claims from opinions and boring statements. However, annotating a sentence as an important or unimportant claim is a non-trivial task as this decision changes depending on who's asking, political context and annotator's background \cite{graves2018understanding}.}

{The model proposed by \citet{hassan2017toward} only learns the labelled instances and does not explore the contextual information of the written text. A context-aware approach in the political discourse environment was introduced by \citet{gencheva-etal-2017-context} who created a rich representation of the sentences from 2016 US presidential debates. Their dataset was compiled by taking the outputs of factchecking of the debates from 9 factchecking organizations. Their models were created to predict if the claim would be highlighted by at least one or by a specific organization. However, the authors don't have any formal definition of claim in their paper, and their model is specific to certain organizations which led to several false positives.}

{Another context-aware approach for claim detection was proposed by \citet{konstantinovskiy2018towards} who used sentence embeddings, pre-trained on a large dataset of NLI. This work also created a crowd-sourced annotated dataset of sentences from UK political TV shows, annotated across 7 classes. However, their classifiers for the fine-grained classification to detect 7 classes of sentences did not yield good results due to lack of enough annotated data, thus requiring more annotations which is a costly and time consuming task.
}

{
We build a model that can be trained in a self-supervised setting to overcome the challenges associated with annotated dataset of claims. We also use attention-based approach to capture \textit{aboutness} and rich contextual information between headline and all the sentences of the article. The performance on manually created test sets demonstrate promising results in identifying ``claim-worthy'' sentences even when no sentence-level annotation was used for training.
}

{\textbf{Fake News Dataset:} A variety of fake news datasets have been released in the recent years, most notably Buzzfeed\footnote{\label{note1}https://www.buzzfeednews.com/article/craigsilverman/these-are-50-of-the-biggest-fake-news-hits-on-facebook-in} and Stanford \cite{allcott2017social} datasets containing list of popular fake news articles from 2016 US presidential elections. However, these datasets only contain webpage URLs of the original article and majority of them don't exist anymore. Following this, several other datasets were published such as Fake news challenge dataset\footnote{http://www.fakenewschallenge.org/} which was used for the task of stance detection; \textit{Getting Real about Fake News} Kaggle dataset\footnote{https://www.kaggle.com/mrisdal/fake-news} which was created by using BS detector tool; and FakeNewsCorpus\footnote{\label{note2}https://github.com/several27/FakeNewsCorpus} which is an open-source large scale collection of fake news articles. However, these articles are labelled as fake based on the domain of the websites they come from. Since, the content of these articles are not verified for degree of veracity, using them directly for training may lead to several false positives.
}\newline
{
\indent This problem was overcome by recently released large dataset, NELA-GT-2018 \cite{nelagt2019}, which contains articles with ground truth ratings retrieved from 8 different assessment sites. However, the label definitions are not generic and dependent on the external organizations. \citet{pathak-srihari-2019-breaking} also introduced intuitive ground truth labels based on the degree of veracity of the fake news articles, however, the dataset is not publicly available. Additionally, they also do not specify the relationship of their labels with the labels used by established fact-checking organizations. Furthermore, due to lack of evidence in these datasets, they cannot be used for downstream task of evidence-based verification, which is one of the motivations of this paper. We overcome all these limitations in our datasets described in the following section. 
}

\section{Datasets}
\label{sec:dataset}
{
We introduce two datasets of compelling fake news articles which have writing style similar to mainstream media. The first dataset, DNF-700, where DNF stands for \textit{\textbf{D}isi\textbf{NF}ormation}, contains articles on politics published within 4 months of 2016 US Presidential Elections from questionable sources (non-mainstream). To compile this dataset, we first extracted fake news articles from working web page URLs of Stanford dataset \cite{allcott2017social}. However, majority of webpage URLs in this dataset are expired and we could extract only 26 fake news articles. Therefore, we then used ``\textit{Getting real about fake news}'' Kaggle\textsuperscript{\ref{note2}} dataset to sample more articles on politics. Since, most of the articles in this dataset contain anomalies (eg: incomplete article, social media comments labelled as fake etc.), we manually verified the writing style and discarded obvious fakes - articles with poor grammar and excessive usage of punctuations. However, the degree of veracity of each article in this dataset is not checked and some articles may contain personal opinions.}\newline
{\indent The second dataset, DNF-300, is more sophisticated subset of DNF-700, containing 290 compelling articles on Politics and 10 on Health/Medical news. Unlike other fake news datasets in which veracity and evidence for articles are not provided, DNF-300 contains articles associated with veracity labels as well as corresponding evidence. The process of annotating this dataset involves identifying sentences from each article based on their persuasive tone and relevance with the headline. These sentences were then queried on the web and top 10 results were considered to gather evidence from credible sources\footnote{Credible sources were extracted from https://mediabiasfactcheck.com/ The sources range between left, center and right biased news sources}. Based on the evidence found, we label the entire article into four categories: \{(0) \textit{false}; (1) \textit{partial truth}; (2) \textit{opinions stated as fact}; (3) \textit{true}\}. These labels are inspired by \cite{pathak-srihari-2019-breaking}; Table-\ref{labelDesc} shows the description and distribution of these labels while the comparison with two popular fact-checking websites is displayed in Figure-\ref{fig:labelComp}. An example from the dataset is shown in Table-\ref{tab:dnf300_partial}}\newline
{
\begin{table}
\centering\scriptsize
\begin{tabular}{|p{0.5cm}|p{1.2cm}|p{1.3cm}|p{1.5cm}|p{1cm}|}
\hline
\textbf{Label} & \textbf{False} & \textbf{Partial True} & \textbf{Opinion Stated As Fact} & \textbf{True}\\
\hline
\textbf{Desc.} & (i) No evidence could be found, or (ii) found evidence to refute the entire article & Article about true event, however, found evidence refuting some of the claims & Article contain false/manipulated claims, however, it's an opinion article which cannot be labelled as fake & Found evidence supporting the entire article\\
\hline
\textbf{Total} & 126 & 75 & 79 & 20\\
\hline
\end{tabular}
\caption{\label{labelDesc}DNF-300 label description and distribution. Claims here are the sentences manually selected based on their persuasive tone and relevance with the headline. Interestingly, some of the articles, which were labelled as fake in other datasets due to the domain of publishing website, turned out to be true news. 
}
\end{table}
}
{
\begin{figure}[h]
\begin{center}
\includegraphics[width=0.48\textwidth]{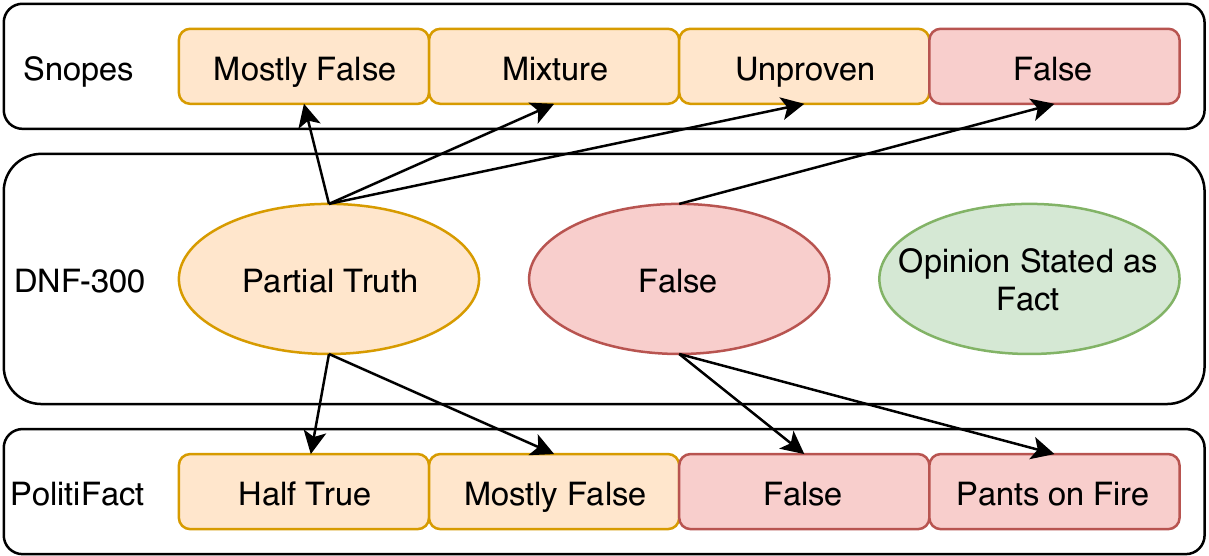}
\caption{\label{fig:labelComp} Label Comparison with Snopes and PolitiFact ratings.}
\end{center}
\end{figure}
}

\begin{table*}[t]
\centering
\begin{tabular}{p{15cm}}
\hline
\textbf{Headline}: Allergens in Vaccines Are Causing Life-Threatening Food Allergies\\
\hline 
It would probably surprise few people to hear that food allergies are increasingly common in U.S. children and around the world . According to one public health website , food allergies in children aged 0-17 in the U.S. increased by 50\% from 1997 to 2011. 
Although food allergies are now so widespread as to have become almost normalized, it is important to realize that millions of American children and adults suffer from severe rapid-onset allergic reactions that can be life-threatening. Foods represent the most common cause of anaphylaxis among children and adolescents. The United Kingdom has witnessed a 700\% increase in hospital admissions for anaphylaxis and a 500\% increase in admissions for food allergy since 1990. 
The question that few are asking is why life-threatening food allergies have become so alarmingly pervasive. A 2015 open access case report by Vinu Arumugham in the Journal of Developing Drugs , entitled `` Evidence that Food Proteins in Vaccines Cause the Development of Food Allergies and Its Implications for Vaccine Policy ,'' persuasively argues that allergens in vaccines—and specifically food proteins—may be the elephant in the room. 
As Arumugham points out, scientists have known for over 100 years that injecting proteins into humans or animals causes immune system sensitization to those proteins. And, since the 1940s, researchers have confirmed that food proteins in vaccines can induce allergy in vaccine recipients. Arumugham is not the first to bring the vaccine-allergy link to the public’s attention. Heather Fraser makes a powerful case for the role of vaccines in precipitating peanut allergies in her 2011 book, The Peanut Allergy Epidemic: What’s Causing It and How to Stop It.\\
\hline
\textbf{Type}: 1 (\textit{Partial Truth})\\
\textbf{Authors}:Admin - Orissa\\
\textbf{URLs}: galacticconnection.com\\
\hline
\textbf{Evidence}: https://www.ncbi.nlm.nih.gov/pmc/articles/PMC3890451/\\
\hline
\textbf{Reason}: The key claim is written in such a way so that it misleads people in thinking all the food related allergies in US are caused by vaccines. Found evidence which says these type of allergies are rare.\\
\hline
\end{tabular}
\caption{\label{tab:dnf300_partial}An example on \textit{Partial Truth} type from DNF-300 dataset. 
}
\end{table*}
{\indent This dataset is also a key contribution of this paper as the articles are manually read and verified. Additionally, the dataset contains two novel features which are essential for the fake news verification task: (i) generic veracity-based label set, independent of any external organization, and (ii) ground truth evidence corresponding to each label. 
}\newline
{
\indent In addition to these two datasets, we also train our model for claim identification on the dataset introduced for context dependent claim detection (CDCD) by \citet{levy_context_2014}. Although this dataset (CDC) does not contain fake news articles, it has manually annotated sentences based on their relevance to a certain topic. These annotations were utilized for the evaluation of our self-supervised learning model described in the following section. More details on the datasets and examples can be found in Appendix.
}
\section{Architecture}
\label{sec:model}
\textbf{Problem Definition:} Given an article with a set of sentences $S=\{S_1, S_2,...S_i,...S_n\}$ and a headline $H$, the task of our multihead attention claim identification network (MA-CIN) is to extract the sentence most relevant to the headline. Our self-supervised model exploits the rich contextual information to extract the relevant sentences which are considered as ``claim-worthy''. \newline
\textbf{Approach:} For this task, we implement two types of attention: (i) self-attention on all sentence vectors so that each sentence $S_i$ is aware of all other sentences in $S$; (ii) cross-attention of headline vector on each sentence vector, so that all self-attended sentences are also aware of the headline's context. We then generate headline based on the context-aware sentences, and compare it with the original headline in three different settings as listed below:
\begin{enumerate}[nolistsep]
\item{\textbf{Headline Vector (MA-CIN (HV)):}} In this setting, the original headline vector acts as the supervisory signal for self-supervised learning. We minimize the mean squared error (MSE) between the generated and the original headline vectors for training.
\item{\textbf{Headline One-Hot Word Vector (MA-CIN (OHWV)):}} In this setting, the words in the original headline act as the supervisory signal. We use LSTM \cite{doi:10.1162/neco.1997.9.8.1735} to predict at most 50 words, from a vocabulary of 20,000 words, to generate a one-hot-vector for each word of the new headline. We then minimize the categorical cross-entropy error (CCE) at each time step corresponding to each word in original and new headlines for training.
\item{\textbf{Combined HV \& OHWV (MA-CIN (Combined)):}} In this setting, both original headline vector and the words act as supervisory signal. Therefore, we combine the two loss functions mentioned above to train the model.
\end{enumerate}
For this, we build several layers in our architecture (see Figure-2), which are delineated as follows:
\begin{figure*}[ht]
\begin{center}
\includegraphics[width=0.8\textwidth]{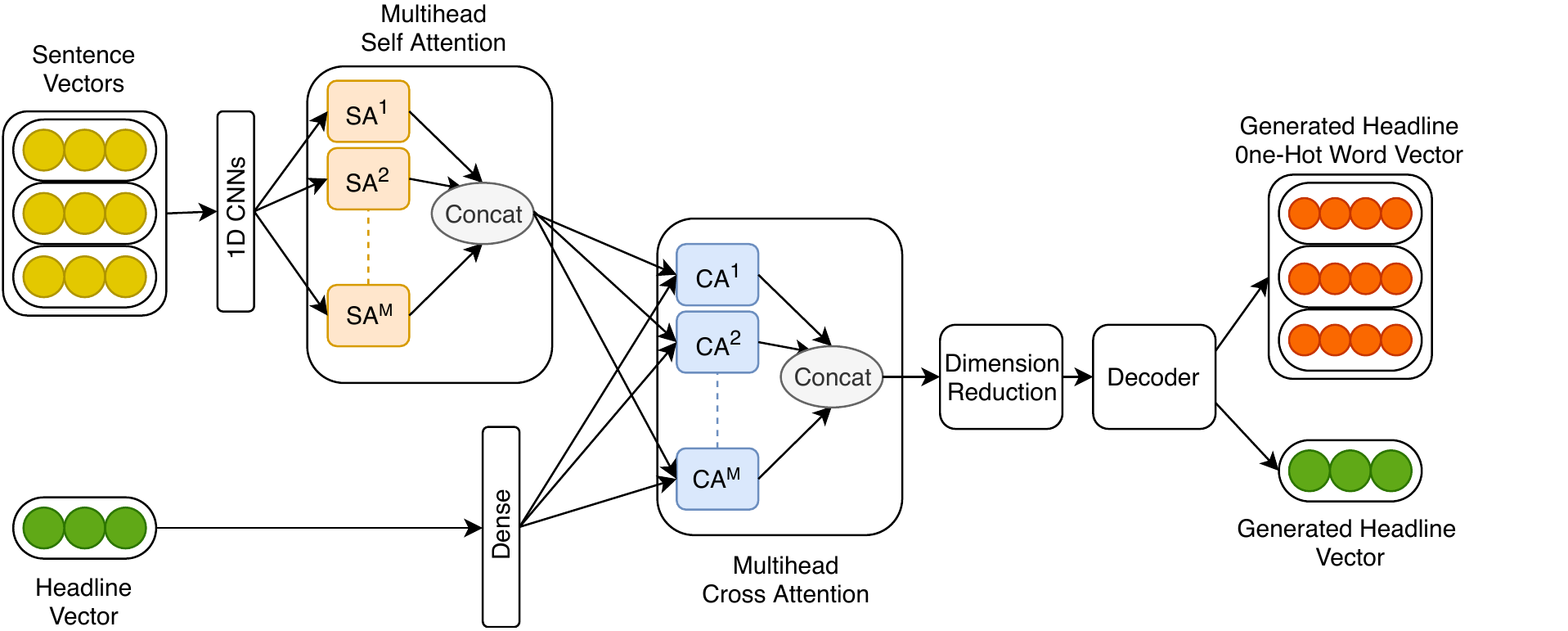}
\caption{\label{fig:architecture}Architecture of Multihead Attention - Claim Identification Network (MA-CIN). The model is trained by using self-supervised learning approach using three variants of supervisory-signal - headline vector, headline words and the combination of both vector and words.}
\end{center}
\end{figure*}
\subsection{Sentence Embeddings}
Each sentence $S_i \in S$, and headline $H$ are converted to a fixed length 300-dimensional vector, $s_i$ and $h$, such that $s_{i}, h \in \mathbb{R}^{1 \times D}$, where $D = 300$. For uniformity, we calculate the maximum number of sentences $L$ that an article can contain in the respective corpus. Next, we zero pad the difference in the quantity of sentence vectors in each article such that every article can be represented as a vector $A \in \mathbb{R}^{L \times D}$.

\subsection{1D Convolution}
To effectively capture local relevance, we leverage 1D-CNN \cite{lecun1998gradient} to extract the features from the article vector $A$. For our experiments the kernel size for each convolution layer is $K \times D \times C$, where $K$ is kernel-width and $C$ is the number of filters. This means the network will process K sentences at a time. The size of K and C is a hyper-parameter and as per our experiments, we set $K=4$ with an assumption that not more than 4 consecutive sentences will be relevant to each other.

\subsection{Self Attention}
\label{subsection:self_attention}
Inspired by the attention implementation in \cite{zhang2018selfattention, vaswani_attention_2017}, to capture global relevance, the article features from the previous 1D-CNN layer are transformed into feature spaces $q$, $k$ to calculate attention, where $q(x) = W_qx$ and $k(x) = W_kx$.

\begin{equation}
    \beta_{j,i} = \frac{exp(z_{ij})}{\sum_{i=1}^{N} exp(z_{ij})}, \text{where } z_{ij} = q(x_{i})^{T}k(x_{j})
    \label{eq:self_attention}
\end{equation}
$\beta \in \mathbb{R}^{N\times{N}}$ is the attention coefficient, which is the normalized relevance score between the sentence $x_{i}$ and $x_{j}$. $\beta$ is then matrix multiplied by $v$, where $v(x) = W_{v}x$, to obtain the context rich output $o_j \in \mathbb{R}^{C \times 1}$. 
\begin{equation}
    o_{j} = \sum_{i=i}^{N} \beta_{j,i}v(x_{i}), \text{where } o_{j} \in \{o_{1}, o_{2}, \dots, o_{N}\}
    \label{eq:context_vector}
\end{equation}
Finally, the output of the self-attention layer is $o \in \mathbb{R}^{C \times N}$, which is computed as
\begin{equation}
    o_j = g(o_j), \text{where, } g(x) = W_{g}x
    \label{eq:output}
\end{equation}
In the above equations, $x \in \mathbb{R}^{C \times N}$ is obtained after applying 1D convolution on sentence vectors, $W_q \in \mathbb{R}^{\overline{C} \times C}$, $W_k \in \mathbb{R}^{\overline{C} \times C}$, $W_v \in \mathbb{R}^{\overline{C} \times C}$, $W_g \in \mathbb{R}^{C \times \overline{C}}$ and output $o \in \mathbb{R}^{C \times N}$. Following the work by \cite{zhang2018selfattention} we preferred the value of $\overline{C}=\frac{C}{8}$ for computation effectiveness. 
We also multiply a $\lambda$ and $\gamma$, learnable scale parameters, to the output of our attention module and input vector respectively to allow the network to choose between local and global sentences effectively.
\begin{equation}
    o = \gamma x + \lambda o
    \label{eq:output_scale}
\end{equation}
$\gamma$ is initialized to 1 and $\lambda$ is initialized to 0, so as to allow the local context to be captured effectively during the early iterations and as the value of $\lambda$ increases it allows the network to add more context to the representation.
\subsection{Multihead Concatenation}
In the architecture, we could apply self attention to the input $x$ M times resulting into M attention heads. The output of one attention head is denoted by $o$. We concatenate the outputs $o^{M}$ to get a richer representation allowing the network to capture various relationships.
\begin{equation}
    msa\_o = \big\Vert_{i=1}^M o^i = o^1 \Vert \dots \Vert o^M
    \label{eq:concat_attention}
\end{equation}
where, $msa\_o \in \mathbb{R}^{MC \times N}$ is the long range context aware output of multihead self attention. Here, \big\Vert \ denotes concatenation across axis $C$.

\subsection{Cross Attention}
The headline vector is transformed into a feature space $\overline{h} = W_hh$, where $\overline{h} \in \mathbb{R}^{\overline{C}\times 1}$ and then, it's relevance is calculated with $msa\_o$, obtained from the previous layer, by using equations defined in \ref{subsection:self_attention}. Finally, after applying multihead concatenation using \ref{eq:concat_attention}, we obtain headline-context aware representation, $mca\_o \in \mathbb{R}^{MC \times N}$. We fix $M=4$ for all our experiments.



\subsection{Loss Function}
To generate the headline vector $d_h$ as close to the input headline vector $h$, we apply Mean Squared Error between $d_h$ and $h$ and calculate the headline vector generation loss $L_{v}$
\begin{equation}
    L_{v} = \frac{1}{n}(\sum^{n}_{i=1} (d_{h_{i}} - h_i )^{2})
    \label{eq:mse}
\end{equation}
For estimating the probability of a word from the vocab in the predicted headline we calculate the cross-entropy between the predicted headline words $d_{hw}$ and input headline one-hot vector $HW$.
\begin{equation}
    L_{w}= - \sum_{i} d_{hw_i} \log (HW_{i})
    \label{eq:cce}
\end{equation}
The total loss $L_{total} = L_v + L_w$ is then evaluated for all samples $b \in B$, where $B$ is one batch.

\begin{table*}[ht]
\begin{center}\scriptsize
\begin{tabular}{cc@{\qquad}ccc@{\qquad}ccc}
  \toprule
  \multirow{2}{*}{\raisebox{-\heavyrulewidth}{Dataset}} & \multirow{2}{*}{\raisebox{-\heavyrulewidth}{Configuration}} & \multicolumn{3}{c}{CDC\_Eval} & \multicolumn{3}{c}{DNF\_Eval} \\
  \cmidrule{3-8}
  & & Prec. & Rec. & F1 & Prec. & Rec. & F1 \\
  \midrule
  $Spacy$ & Baseline & 0.09 & 0.14 & 0.11 & 0.33 & 0.42 & 0.37 \\
  \midrule
  $CDC$ & Baseline \cite{levy_context_2014} & 0.23 & - & - & - & - & - \\
   & MA-CIN(HV) & 0.18 & 0.08 & 0.11 & 0.39 & 0.53 & 0.45 \\
    & MA-CIN(OHWV) & 0.25 & 0.10 & 0.15 & 0.40 & 0.54 & 0.46 \\
    & MA-CIN(Combined) & \textbf{0.26} & \textbf{0.11} & \textbf{0.16} & \textbf{0.42} & \textbf{0.57} & \textbf{0.48} \\
  \midrule
  $DNF-700$ & MA-CIN(HV) & 0.20 & 0.09 & 0.12 & 0.37 & 0.54 & 0.44 \\
    & MA-CIN(OHWV) & 0.19 & 0.08 & 0.11 & 0.40 & 0.5 & 0.44 \\
    & MA-CIN(Combined) & \textbf{0.28} & \textbf{0.12} & \textbf{0.16} & \textbf{0.41} & \textbf{0.55} & \textbf{0.47} \\
    
  \midrule
  $DNF-300$ & MA-CIN(HV) & 0.19 & 0.08 & 0.11 & 0.39 & 0.53 & 0.48 \\
    & MA-CIN(OHWV) & \textbf{0.25} & \textbf{0.11} & \textbf{0.15} & 0.38 & 0.53 & 0.45 \\
    & MA-CIN(Combined) & 0.24 & 0.10 & 0.14 & \textbf{0.42} & \textbf{0.57} & \textbf{0.48} \\
  \bottomrule
\end{tabular}
\end{center}

\caption{\label{tab:results} Comparison of MA-CIN model configurations over three datasets and two evaluation sets for identification of ``claim-worthy'' sentences.}
\end{table*}

\section{Experiments and Evaluation}
\label{sec:exp}

\begin{figure*}[ht]
\begin{center}
\includegraphics[width=1.0\textwidth]{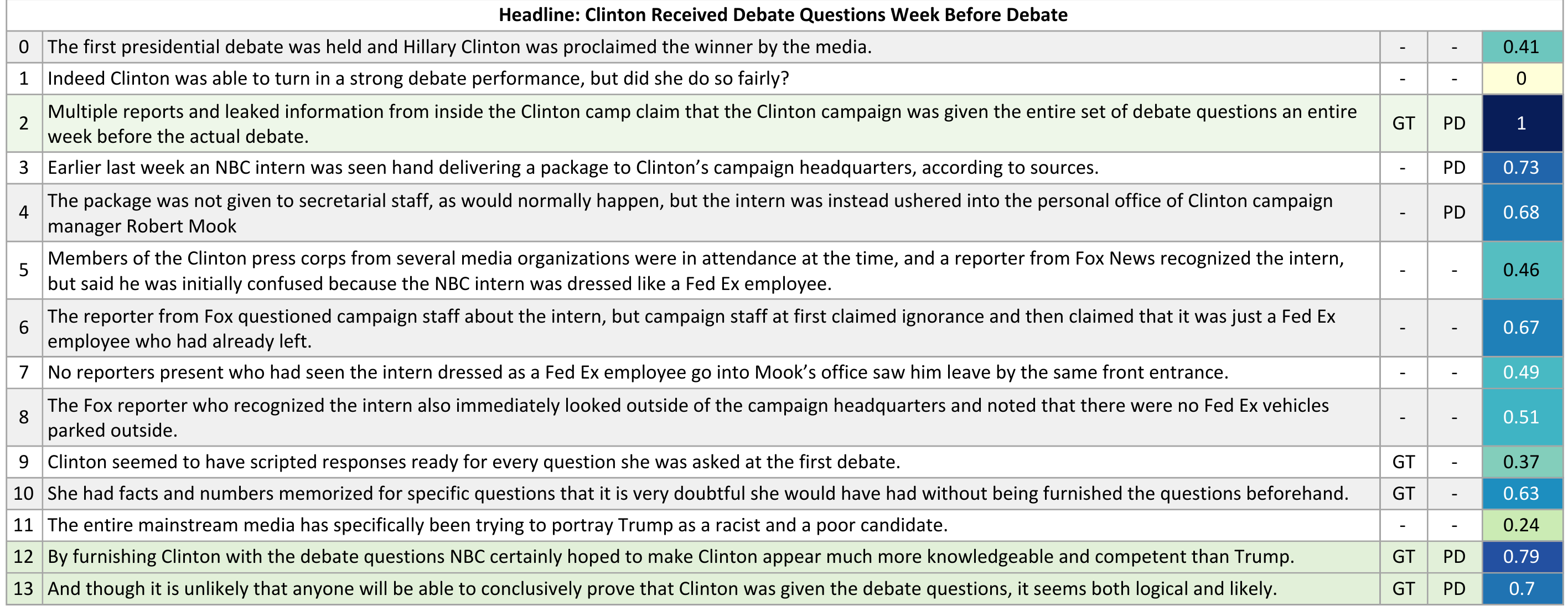}
\caption{\label{fig:hd_sent} Interpretation of relevance of sentences with the headline of an example article from DNF-300. GT and PD indicate ground truth and top-5 predicted ``claim-worthy'' sentences, respectively. MA-CIN model was able to predict 3 most relevant sentences correctly. Last column shows the attention weights between headline and each of the sentences of the article. Sentence 2 has been correctly predicted as the most relevant while sentence 1 is the least relevant.}
\end{center}
\end{figure*}

\subsection{Training Setup}
\label{sect:exp_training}
We train our Multihead Attention model for Claim Identification, MA-CIN, on datasets mentioned in Section \ref{sec:dataset}. The CDC dataset contains total of 522 articles. Amongst these, there are 47 articles with 8 or more annotated claim sentences which are considered as evaluation set (CDC\_Eval) for this dataset. Next, for DNF-300 and DNF-700, we asked two annotators to manually tag at least 5 sentences as ``claim-worthy'' in each of the 50 articles. Sentences which were consented by both the annotators as ``claim-worthy'' were finalized as ground truth claims for these 50 articles, and used as testing set for evaluating the model performance on DNF datasets. The remaining 475 articles from CDC, 250 articles from DNF-300, and 650 articles from DNF-700 were split into 5 folds to train the model using a 5-Fold cross validation \cite{Kohavi:1995:SCB:1643031.1643047}, where we use 4 folds for training and 1 fold for validation. Each of the three settings, described in Section- \ref{sec:model}: MA-CIN(HV), MA-CIN(OHWV) and MA-CIN(Combined), was trained with each of the three datasets, and evaluated on DNF\_Eval and CDC\_Eval. Total number of parameters for these three settings are 15,012,916 (10,240 non-trainable), 40,975,656 (10,240 non-trainable) and 41,812,564 (12,288 non-trainable) respectively. All other network parameters are displayed in supplemental material.\newline
\indent In each setting, we use batch normalization, ReLU non-linearity as an activation function, and a dropout of 0.5 for every convolution operation. We trained all the models for 2000 epochs, where, for every training we used Adam optimizer with a learning rate $lr=0.0001$, $\beta_{1}=0.99$ and $\beta_{2}=0.0$. There was no weight decay set as the model was trained in a self-supervised setting with finite epochs and an already small learning rate. Glove 300D word embedding was used for all our experiments and the number of input sentences was set to 500. The models were trained on three 11GiB Nvidia 1080Ti GPUs in parallel.

\subsection{Evaluation Metrics}
\label{sect:eval_metrics}
We evaluate MA-CIN models on two evaluation sets, DNF\_Eval and CDC\_Eval. With self-supervised setting we first rank the sentences based on relevance with the headline and then extract the top five sentences along with their sentence ids as ``claim-worthy'' sentences. For evaluation on DNF\_Eval, we calculate the \textit{true positives}(TP), \textit{false positives}(FP) and \textit{false negatives}(FN) from ground truth claim ids. To evaluate on CDC\_Eval, since we do not have ground truth claim ids, we calculate cosine similarity between the extracted sentences and the ground truth claims. We experiment with various similarity threshold to calculate TP, FP and FN, and set the final threshold to 0.95 to report best performing results. Finally, these metrics are used to report Precision@1, Recall@1 and F-1 scores.

\subsection{Results}
\label{sect:results}
Table-\ref{tab:results} shows the performance of baseline (CDC) \citep{levy_context_2014} and three variants of MA-CIN models. We report two baselines - (1) spacy, and (2) \citet{levy_context_2014} using supervised learning method on CDC dataset which contains annotated claims. Since, \citet{levy_context_2014} do not report Recall and F1 scores, we have reported their Precision@1 score in this paper. We also train MA-CIN models on this dataset by removing all the annotations for self-supervised training. We observe that: 
\begin{enumerate}[nolistsep]
    \item The combined variant of our self-supervised approach performs slightly better than the baseline on the CDC dataset. This shows that, MA-CIN models are able to learn similar properties as the baseline but without any sentence-level annotations. Thus, this eliminates the need to have an annotated dataset for claim identification.
	\item MA-CIN models give comparable results on all three datasets. This shows the scalability of the models to identify ``claim-worthy'' sentences from any given article.
	\item The combined variant of MA-CIN, which generates both the headline vector and the word in headline, performs better on all the datasets, except one: MA-CIN (OHWV) model trained on DNF-300 and evaluated on CDC\_Eval performs slightly better than the combined model, however, the difference in the performance is very small. 
\end{enumerate}

\begin{figure}[ht]
\begin{center}
\includegraphics[width=0.5\textwidth]{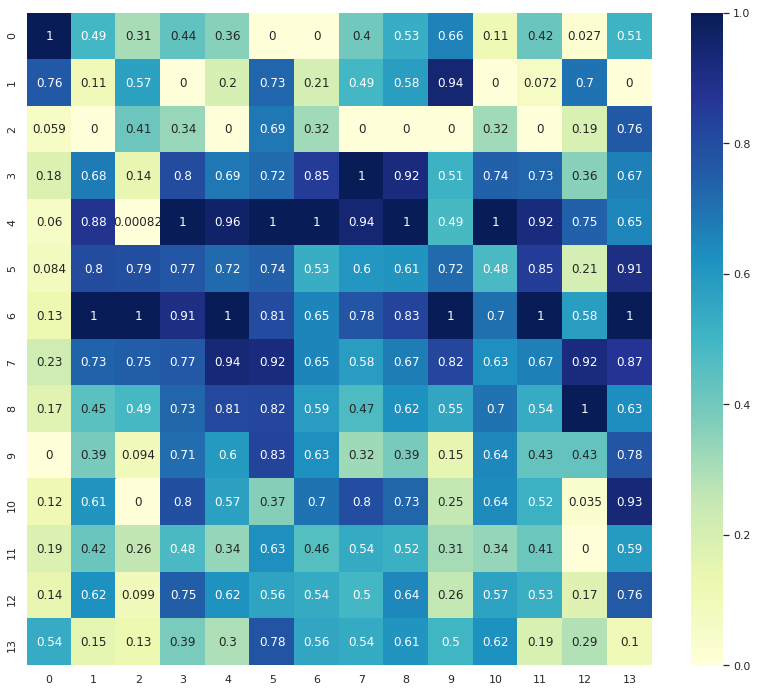}
\caption{\label{fig:sent_sent}Interpretation of sentence-to-sentence relevance through attention weights.}
\end{center}
\end{figure}
\addtolength{\textfloatsep}{-0.30in}

\section{Discussion}
\label{sec:discuss}

\subsection{Analyzing Attention Weights}
\label{sect:analysis}
Attention weights help make the model interpretable to the end users by depicting relationship between all sentences as well as with the headline. From Figure-\ref{fig:hd_sent}, we can see that out of the top-5 predicted claims, 3 of them are present in the human evaluated test set. The last column, which contains attention coefficients between the headline and each sentence, depicts some interesting results - 

(i) based on the human evaluation, the sentence having the least relevance with the headline is sentence 1. While this sentence contains words also present in the headline, the underlying meaning is not the same. This has been successfully captured by MA-CIN model by predicting sentence 1 as the least relevant claim;

(ii) further, highly ranked sentences 2, 12, and 13 have been correctly predicted as relevant claims by the model. This shows the model's ability in learning the semantic relationship between the headline and the content of the article, and subsequently putting importance on sentences that are relevant to the headline's underlying meaning. This property, which is also called ``\textit{aboutness}", is efficiently exhibited by the model.

(iii) sentence 3, which is predicted by MA-CIN model as relevant with a score of 0.73, is not present in the ground truth. This indicates that the two annotators did not agree to have this sentence verified, even if it is relevant to the point of the article. To analyze this further, we plan to conduct user studies as one of the future avenues.

(iv) sentence 4 is also predicted as a relevant claim but it's missing from the ground truth since the annotators did not agree to have this verified. The reason for this prediction could be because self-attention is able to identify the premise of highly relevant sentences. Hence, sentence 4, which is the continuation of highly relevant sentence 3, is also given importance by the headline. This relevance between sentences 3 and 4 is depicted in Figure-\ref{fig:sent_sent}, where the attention weight between these two is the highest. 

From Figure-\ref{fig:sent_sent}, we also observe that:

(i) sentence 4 is highly relevant to sentences 3 to 8, which is intuitive, since the story of the intern forms the premise of the claims in the article;

(ii) sentences 2 and 4 have been shown to have the least relevance with each other which is also true as shown in Figure-\ref{fig:hd_sent}. The two sentences, if considered in isolation, make two different claims which are not related to each other;

(iii) the model has made sure that a sentence does not assign high relevance to itself as it would be counter-intuitive.

\subsection{Limitations}
\label{sect:limitation}
Since, the evaluation methodologies for CDC dataset has not been explained clearly, in our paper, we have considered vector cosine similarity between the ground truth claim in the CDC\_Eval and the extracted claim from the model which may leave a margin of error in the evaluation scores. Additionally, ground truth in DNF\_Eval is manually generated and may contain subjective biases. Although these biases have been overcome by MA-CIN models, as explained in \ref{sect:analysis}, but we also plan to enhance the ground truth judgement using crowdsourced annotation. We intend to use these annotations to fine-tune the models.

\section{Conclusion and Future Work}
{In this work, we build a novel, self-supervised approach to identify ``claim-worthy'' sentences - an important task for automated fact checking. The focus of this work is on fake news articles where there is a deliberate intent to influence people or cause social unrest. We have introduced novel datasets of such articles with features essential for the downstream task of fake news verification. Using powerful attention models, we explore the notion of \textit{aboutness} of the headline and the content of the article to identify ``claim-worthy'' sentences. Experiments with three datasets show the strength of our model architecture in overcoming human-induced biases, which is quite common when using sentence-level claim-annotated datasets. Based on the comparison with the baseline, which was implemented using annotated dataset, we show that our models do not require annotated claims for training to identify claim-worthy sentences efficiently. We have also showed that our model is scalable to any dataset with topic and content.}\newline
{\indent Future work involves increasing the robustness of the models presented in this paper. We plan to use crowdsourced annotation on the dataset released with this paper to measure the \textit{influence} of the article on general readers and then use these indicators to fine-tune our models. Experimentation with more robust sentence encoders is another avenue of future work. Additionally, going forward, we plan to identify a maximum of 3 claims per article which will be used for evidence-based fake news detection. We also plan to expand the dataset, presented in this work, to include fake news articles on topics other than Politics and Health. 
}

\bibliography{acl2020}
\bibliographystyle{acl_natbib}

\appendix

\section{Appendices}
\label{sec:appendix}
\subsection{Definitions}
\label{sect:definition}
\textbf{Fake News:} Articles where there is a deliberate intent to influence readers through fabricated or manipulated claims in the headline and the content. Such articles have a compelling writing style similar to the mainstream media. \newline
\textbf{``Claim-worthy'':} Statements which are important to the point of the article but one would require to have them verified. \newline
\textbf{Compelling Fake News Articles:} Articles which make persuasive claims in headline and content, that may influence readers to believe a fabricated/manipulated story. \newline 
\textbf{Credible Sources}:  Mainstream media, established fact-checking websites and Government documents. \newline
\textbf{Questionable Sources}:  Non-mainstream media like infowars, naturalnews, breitbart etc.

\subsection{Experiment Architectures}
\subsubsection{Vector Generation}
\begin{figure}[h]
\begin{center}
\includegraphics[width=0.45\textwidth]{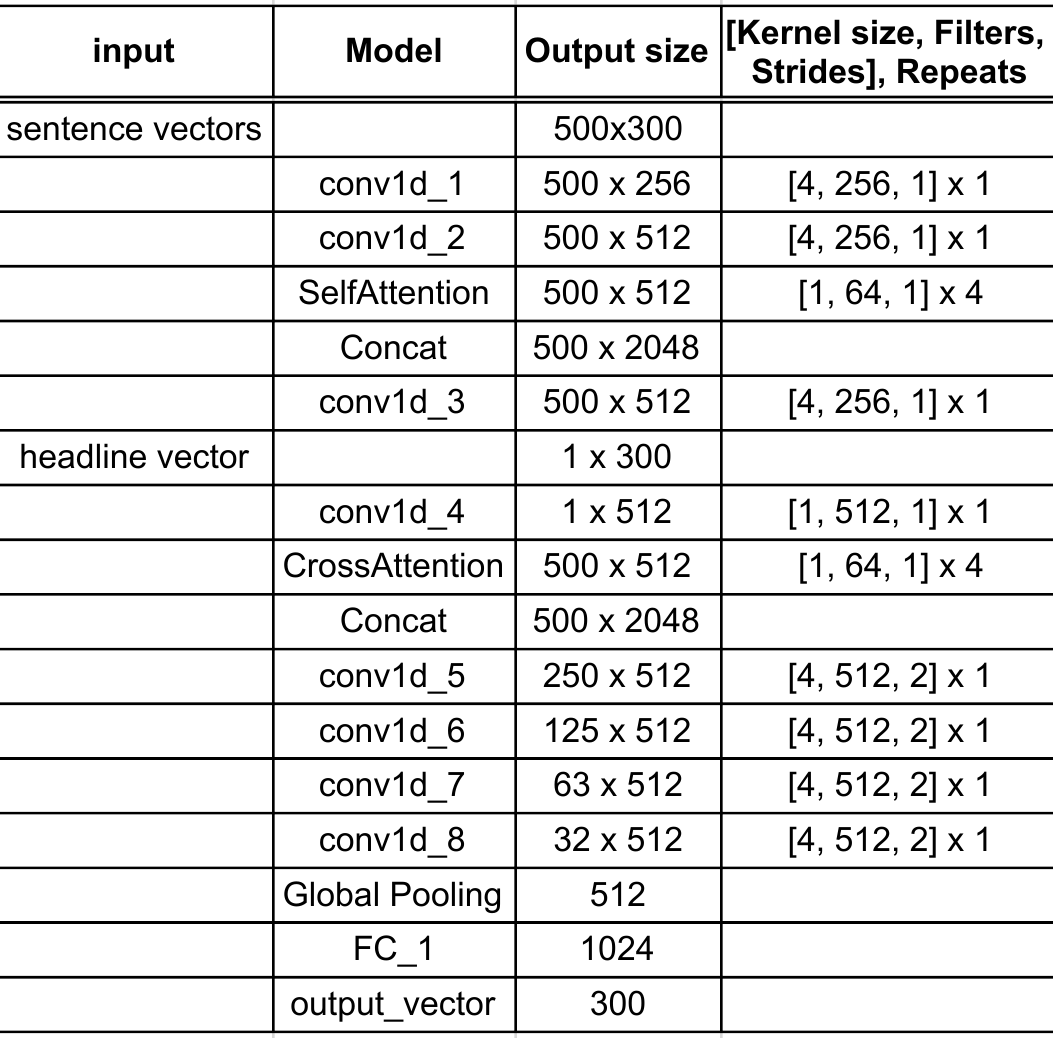}
\caption{\label{fig:arch_setting_hv}Architecture setting for generating Headline Vector(HV).}
\end{center}
\end{figure}

Architecture setting for generating Headline Vector (HV) displayed in Figure-\ref{fig:arch_setting_hv}
\subsubsection{Word Generation}
\begin{figure}[h]
\begin{center}
\includegraphics[width=0.45\textwidth]{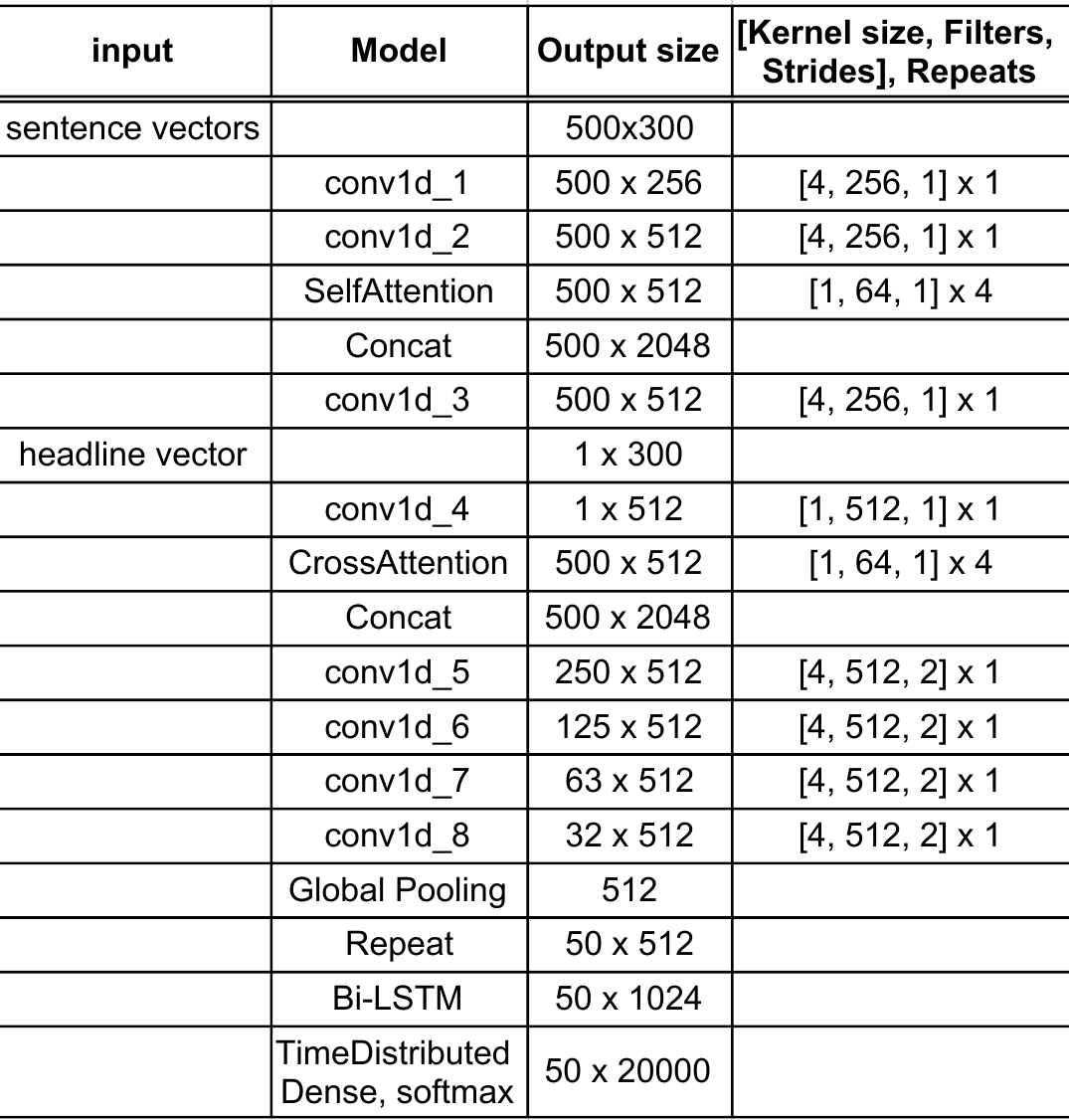}
\caption{\label{fig:arch_setting_ohw}Architecture setting for generating Headline Vector Word Probabilities (OHWV).}
\end{center}
\end{figure}

Architecture setting for generating Headline Vector Word Probabilities (OHWV) displayed in Figure-\ref{fig:arch_setting_ohw}

\subsection{DNF-700 Dataset Details} 
\label{sect:700dataset}
Each article is identified by an \textit{id}. The content of the article is stored in a separate text files having file name {``article\_\textit{id}''}, for example, {article\_122}. A JSON file is also provided with the following fields:\newline \newline
\textbf{id:} Unique identifier of the article starting from 0.\newline
\textbf{authors:} Authors of the article.\newline
\textbf{headline:} Headline of the article. \newline
\textbf{type:} ``fake'' (articles from Stanford and Buzzfeed datasets which are already proven fake); and ``questionable'' (articles from \textit{Getting Real About Fake News} Kaggle dataset which require manual verification of the degree of veracity)  \newline
\textbf{urls:} Source/domain URL of the article. \newline

\subsection{DNF-300 Dataset Details} 
\label{sect:300dataset}
DNF-300 is more sophisticated subset of DNF-700 with additional fields based on manual verification of the article. The JSON file of this dataset contains following fields:\newline \newline
\textbf{id:} Unique identifier of the article starting from 0.\newline
\textbf{authors:} Authors of the article.\newline
\textbf{headline:} Headline of the article. \newline
\textbf{type:} \{(0) False; (1) Partial Truth; (2) Opinions Stated As Fact; (3) True\}  \newline
\textbf{urls:} Source/domain URL of the article. \newline
\textbf{evidence:} URL of credible sources supporting or refuting the article. This field is empty when no evidence were found which talked about the claims made in this article. This means, the claims are innovated lies. In such cases, the \textit{type} field is set as 0. \newline
\textbf{reason:} Reason about the verdict. It can be one of the following:
\begin{enumerate}[nolistsep]
	\item Based on Snopes rating 'False' which means 'the primary elements of a claim are demonstrably false.'  
	\item Based on Snopes rating 'Unproven' which means 'insufficient evidence exists to establish the given claim as true, but the claim cannot be definitively proved false.'
	\item Based on Snopes rating 'Mixture' which means 'a claim has significant elements of both truth and falsity to it such that it could not fairly be described by any other rating.'
	\item Based on Snopes rating 'Mostly False' which means 'the primary elements of a claim are demonstrably false, but some of the ancillary details surrounding the claim may be accurate.'
	\item The key claim is false (based on Snopes rating), however, the article also contains opinions stated as fact.
	\item Snopes mentiones that a true story was manipulated to mislead people.
	\item The key claims are true but exaggerated by adding personal opinions stated as fact.
	\item No reports from trusted sources for the key claims.
	\item True story manipulated to mislead readers by making unverifiable claims such as \textit{`some\_claim'}.
	\item Article is fraught with opinions stated as fact about a true event.
	\item Found evidence to refute key claims.
	\item Article contains opinions stated as fact.
	\item Evidence found to support key claims.
\end{enumerate}

\begin{figure}[ht]
\begin{center}
\includegraphics[width=0.5\textwidth]{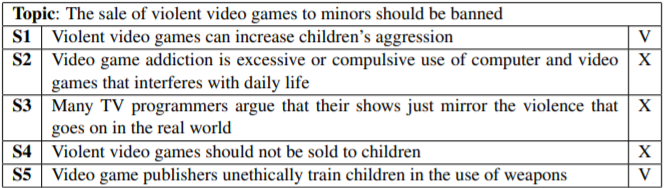}
\caption{\label{fig:cdc_eg}: Example for CDCs and for statements that should not be considered as CDCs. The V and X indicate if the candidate is a CDC for the given Topic, or not, respectively.}
\end{center}
\end{figure}

\subsection{Examples}
\label{sect:example}
We present examples for all 4 label types \{\textit{False}; \textit{Partial Truth}; \textit{Opinion stated as fact}; \textit{True}\} present in our dataset: DNF-300. Please refer Table-\ref{tab:dnf300_false},\ref{tab:appendnf300_partial},\ref{tab:dnf300_opinion},\ref{tab:dnf300_true}. An annotated example from CDC dataset is displayed in Figure-\ref{fig:cdc_eg}

\begin{table*}
\centering
\begin{tabular}{p{15cm}}
\hline
\textbf{Headline}: Clinton Received Debate Questions Week Before Debate\\
\hline 
The first presidential debate was held and Hillary Clinton was proclaimed the winner by the media. Indeed Clinton was able to turn in a strong debate performance, but did she do so fairly? Multiple reports and leaked information from inside the Clinton camp claim that the Clinton campaign was given the entire set of debate questions an entire week before the actual debate. Earlier last week an NBC intern was seen hand delivering a package to Clinton’s campaign headquarters, according to sources. The package was not given to secretarial staff, as would normally happen, but the intern was instead ushered into the personal office of Clinton campaign manager Robert Mook. Members of the Clinton press corps from several media organizations were in attendance at the time, and a reporter from Fox News recognized the intern, but said he was initially confused because the NBC intern was dressed like a Fed Ex employee. The reporter from Fox questioned campaign staff about the intern, but campaign staff at first claimed ignorance and then claimed that it was just a Fed Ex employee who had already left. No reporters present who had seen the intern dressed as a Fed Ex employee go into Mook’s office saw him leave by the same front entrance. The Fox reporter who recognized the intern also immediately looked outside of the campaign headquarters and noted that there were no Fed Ex vehicles parked outside. Clinton seemed to have scripted responses ready for every question she was asked at the first debate. She had facts and numbers memorized for specific questions that it is very doubtful she would have had without being furnished the questions beforehand. The entire mainstream media has specifically been trying to portray Trump as a racist and a poor candidate. By furnishing Clinton with the debate questions NBC certainly hoped to make Clinton appear much more knowledgeable and competent than Trump. And though it is unlikely that anyone will be able to conclusively prove that Clinton was given the debate questions, it seems both logical and likely.\\
\hline
\textbf{Type}: 0 (\textit{False})\\
\textbf{Authors}:Baltimore Gazette\\
\textbf{URLs}: http://www.freemarketcentral.com/index.php/post/2503/report-clinton-received-debate-questions-a-week-before-debate\\
\hline
\textbf{Evidence}: [https://www.snopes.com/fact-check/clinton-received-debate-questions-week-before-debate/, https://www.truthorfiction.com/hillary-clinton-received-debate-questions-advance/]\\
\hline
\textbf{Reason}: Based on Snopes rating 'False' which means 'the primary elements of a claim are demonstrably false.'\\
\hline
\end{tabular}
\caption{\label{tab:dnf300_false}An example on \textit{False} type from DNF-300 dataset. 
}
\end{table*}

\begin{table*}[t]
\centering
\begin{tabular}{p{15cm}}
\hline
\textbf{Headline}: Allergens in Vaccines Are Causing Life-Threatening Food Allergies\\
\hline 
It would probably surprise few people to hear that food allergies are increasingly common in U.S. children and around the world . According to one public health website , food allergies in children aged 0-17 in the U.S. increased by 50\% from 1997 to 2011. 
Although food allergies are now so widespread as to have become almost normalized, it is important to realize that millions of American children and adults suffer from severe rapid-onset allergic reactions that can be life-threatening. Foods represent the most common cause of anaphylaxis among children and adolescents. The United Kingdom has witnessed a 700\% increase in hospital admissions for anaphylaxis and a 500\% increase in admissions for food allergy since 1990. 
The question that few are asking is why life-threatening food allergies have become so alarmingly pervasive. A 2015 open access case report by Vinu Arumugham in the Journal of Developing Drugs , entitled `` Evidence that Food Proteins in Vaccines Cause the Development of Food Allergies and Its Implications for Vaccine Policy ,'' persuasively argues that allergens in vaccines—and specifically food proteins—may be the elephant in the room. 
As Arumugham points out, scientists have known for over 100 years that injecting proteins into humans or animals causes immune system sensitization to those proteins. And, since the 1940s, researchers have confirmed that food proteins in vaccines can induce allergy in vaccine recipients. Arumugham is not the first to bring the vaccine-allergy link to the public’s attention. Heather Fraser makes a powerful case for the role of vaccines in precipitating peanut allergies in her 2011 book, The Peanut Allergy Epidemic: What’s Causing It and How to Stop It.\\
\hline
\textbf{Type}: 1 (\textit{Partial Truth})\\
\textbf{Authors}:Admin - Orissa\\
\textbf{URLs}: galacticconnection.com\\
\hline
\textbf{Evidence}: https://www.ncbi.nlm.nih.gov/pmc/articles/PMC3890451/\\
\hline
\textbf{Reason}: The key claim is written in such a way so that it misleads people in thinking all the food related allergies in US are caused by vaccines. Found evidence which says these type of allergies are rare.\\
\hline
\end{tabular}
\caption{\label{tab:appendnf300_partial}An example on \textit{Partial Truth} type from DNF-300 dataset. 
}
\end{table*}

\begin{table*}[t]
\centering
\begin{tabular}{p{15cm}}
\hline
\textbf{Headline}: George Soros: Trump Will Win Popular Vote by a Landslide but Clinton Victory a 'Done Deal'\\
\hline 
In recent weeks, Democrats have attempted to paint Republican presidential nominee Donald J. Trump as a lunatic for claiming that the election is going to be rigged in favor of his Democratic rival, Hillary Clinton. Even Republican politicians and former politicians are telling Trump to knock off such talk. But, as usual, Trump's shrewdness and defiance of standard political decorum – in which the ``opposition'' party merely rolls over and surrenders in the face of Democratic pressure – is winning the day. None other than billionaire investor and longtime Democratic supporter George Soros has said that the fix is literally in for the election, in favor of Clinton – no matter how much of the popular vote, and from which battleground states, Trump captures. As reported by Top Right News and other outlets, during a recent interview with Bloomberg News, Soros – a Democrat mega-donor – openly admitted that Trump will win the popular vote in a ``landslide.'' However, he said that none of that would matter, because a President Hillary Clinton is already a ``done deal.'' In the interview, which is now going viral, Soros says with certainty that Trump will take the popular vote, despite what the polls say now (which are completely rigged to oversample Democrats), but not the Electoral College, which will go to Clinton. When the reporter asks if that is already a ``done deal'' – that Clinton will be our next president no matter what – Soros says ``yes,'' and nods his head. Is Soros just making a prediction out of overconfidence? Or does he truly know something most of us don't know?\\
\hline
\textbf{Type}: 2 (\textit{Opinion Stated As Fact})\\
\textbf{Authors}:J. D. Heyes\\
\textbf{URLs}: https://www.naturalnews.com/055789\_George\_Soros\_Hillary\_Clinton\_electoral\_college.html\\
\hline
\textbf{Evidence}: 1. https://www.snopes.com/fact-check/george-soros-trump-will-win-popular-vote-by-a-landslide-but-clinton-victory-a-done-deal/,\\ 2. https://www.bloomberg.com/news/videos/2016-01-22/soros-clinton-to-win-popular-vote-in-landslide\\
\hline
\textbf{Reason}: The key claim is false (based on Snopes rating), however, the article also contains opinions stated as fact.\\
\hline
\end{tabular}
\caption{\label{tab:dnf300_opinion}An example on \textit{Opinion stated as fact} type from DNF-300 dataset. 
}
\end{table*}

\begin{table*}[t]
\centering
\begin{tabular}{p{15cm}}
\hline
\textbf{Headline}: Donald Trump: Minnesota Has ‘Suffered Enough’ Accepting Refugees\\
\hline 
In a pitch to suspend the nation’s Syrian refugee program , Donald Trump said Minnesotans have ``suffered enough'' from accepting Somali immigrants into their state. ``Here in Minnesota you have seen first hand the problems caused with faulty refugee vetting, with large numbers of Somali refugees coming into your state, without your knowledge, without your support or approval,'' Trump said at a Minneapolis rally Sunday afternoon. He said his administration would suspend the Syrian refugee program and not resettle refugees anywhere in the United States without support from the communities, while Hillary Clinton’s ``plan will import generations of terrorism, extremism and radicalism into your schools and throughout your communities.''\\
\hline
\textbf{Type}: 3 (\textit{True})\\
\textbf{Authors}:Henry Wolff\\
\textbf{URLs}: amren.com\\
\hline
\textbf{Evidence}: 1. https://time.com/4560078/donald-trump-minnesota-somali-refugees/,\\
2. https://www.buzzfeednews.com/article/claudiakoerner/trump-vs-somali-refugees\\
\hline
\textbf{Reason}: Evidence found to support key claims.\\
\hline
\end{tabular}
\caption{\label{tab:dnf300_true}An example on \textit{True} type from DNF-300 dataset. 
}
\end{table*}



\end{document}